\documentclass[11pt]{article}
\usepackage[utf8]{inputenc}
\usepackage[T1]{fontenc}
\usepackage{lmodern}
\usepackage[margin=1in]{geometry}
\usepackage{microtype}
\usepackage{graphicx}
\usepackage{booktabs}
\usepackage{array}
\usepackage{longtable}
\usepackage{authblk}
\usepackage[authoryear,round]{natbib}
\usepackage[hidelinks]{hyperref}
\graphicspath{{figures/}}

\title{\bfseries Same question, different history: language, national identity, and credit in large language models}
\author[1]{William Guey}
\author[1]{Wei Zhang}
\author[2]{Vitor D. de Moura}
\author[1]{Pierrick Bougault}
\author[3]{Jos\'e O. Gomes}
\affil[1]{Department of Industrial Engineering, Tsinghua University, Beijing, China}
\affil[2]{School of Social Sciences, Tsinghua University, Beijing, China}
\affil[3]{Department of Industrial Engineering, Federal University of Rio de Janeiro, Brazil}
\date{}

\begin{document}
\maketitle

\begin{abstract}
\noindent
Who invented the radio, Russia's Alexander Popov or Italy's Guglielmo Marconi? Was the telephone the achievement of Bell in the United States or Meucci in Italy? Does printing belong to China's Bi Sheng or Germany's Gutenberg? The answer depends not only on the historical record but also on language, memory, and the perspective from which the question is asked. As large language models become a default gateway to information, they increasingly shape how such contested attributions are presented. We analyse eleven widely used models across 21 disputed inventions and discoveries, in twelve languages and across 75,896 responses. While the models generally acknowledge that credit is contested, the language of the question quietly reshapes who is brought forward as the ``inventor'': asked in a claimant's associated national language, that figure is significantly more likely to appear. Russian prompts tend to surface Popov, Chinese prompts raise the visibility of Bi Sheng, and Portuguese prompts strengthen Santos-Dumont, while dominant Anglophone figures remain stable across all languages and are retrieved almost regardless of how one asks. These patterns hold after adjusting for response length, differences between models, the relative prominence of claimants in the historical record, and levels of national commemoration. Language acts as a switch that activates different national versions of the same history: the same question does not return the same past, but returns different national memories depending on the language in which it is asked. Large language models therefore do not simply retrieve knowledge; they help determine which histories become visible and which remain in the background, subtly shaping how collective memory is reproduced in the digital age. We interpret this pattern as a computational form of everyday, or banal, nationalism, in which collective memory is differentially activated across linguistic communities.

\vspace{0.6em}
\end{abstract}

\section{Introduction}

Ask who invented the radio and the answer depends on where you stand. In Italy it is Guglielmo Marconi; in Russia it is Alexander Popov, honored each year on a national Radio Day. Ask who invented the telephone and Americans will say Alexander Graham Bell, while Italians point to Antonio Meucci, whose claim the United States House of Representatives formally honored in 2002 \citep{hres269meucci2002}. Ask who invented printing with movable type and the German answer, Johannes Gutenberg, competes with the Chinese answer, Bi Sheng, who worked four centuries earlier. These are not trivia with hidden correct answers. They are contested attributions, and which claimant a culture credits is a social fact about that culture rather than a discovery about the past.

Sociologists of science have long understood this. Robert Merton observed that the same discovery is frequently made independently by several people at once, a pattern he called multiple discovery, so that priority disputes are a structural feature of science rather than an accident \citep{merton1961multiples,merton1957priorities}. Stephen Stigler distilled a related irony into his law of eponymy: discoveries are routinely named after someone other than the person who made them first \citep{stigler1980eponymy}. If credit and priority come apart so often, then the act of naming an inventor is a choice, and choices of this kind are made by societies for reasons of their own.

Nations make such choices deliberately and at scale. The historian David Edgerton describes techno-nationalism, the habit of claiming technologies as proof of a nation's distinctive genius, as a pervasive feature of modern political culture \citep{edgerton2007technonationalism}. Credit, once claimed, is then fixed in place through commemoration. Here the paper draws on the tradition of memory studies. Maurice Halbwachs established that remembering is itself a collective act, structured by the social frameworks of the present rather than a neutral recording of the past \citep{halbwachs1992collective}, and later scholars showed how such cultural memory is bound up with the identity of the group that maintains it \citep{assmann1995collective}. Building on this, Pierre Nora's concept of lieux de m\'emoire, or sites of memory, captures how monuments, museums, banknotes, anniversaries, and named institutions congeal a shared past into an official version that citizens absorb without argument \citep{nora1989memory}. Such commemoration is politically constructed and contested, built to serve national identity rather than to settle a historical record \citep{gillis1994commemorations}, and it is reinforced in the school curricula and textbooks through which each nation selects and celebrates its heroes \citep{hutchins2016nationalism}. The Soviet campaign that made Popov the inventor of radio, complete with an annual holiday, is a textbook case of commemoration manufacturing credit \citep{susskind1962radio}. Crucially, this memory work is uneven across nations. The narratives of powerful, globally dominant cultures circulate widely, while the claims of smaller or historically subordinated nations remain visible mainly to their own publics, a long-running asymmetry in whose knowledge counts as universal.

This is the terrain of nationalism studies. Nations are, in Benedict Anderson's influential account, imagined communities, held together less by face-to-face contact than by shared media that let dispersed strangers picture themselves as one people \citep{anderson1991imagined}; print was the original such medium, and the heritage each community celebrates is itself in part an invented and institutionalized tradition \citep{hobsbawm1983invention}. Nationhood is reproduced not only in the conspicuous moments of flags and anthems but in the unwaved, everyday background that Michael Billig calls banal nationalism, the routine and barely noticed flagging of the homeland in ordinary discourse \citep{billig1995banal}. On this understanding nationhood is best treated not as a fixed essence but as a practical category, continually enacted and reactivated in everyday classification and talk \citep{brubaker1996nationalism}. A claimant's place in a national story is therefore reproduced wherever that story is routinely retold, and the medium of retelling matters.

Increasingly, the place people take these questions is not a library or a classroom but a chatbot. When a large language model, the statistical text-prediction system behind tools such as ChatGPT, answers ``who invented the radio,'' it joins this lineage of everyday media through which nationhood is reproduced, performing the same act of credit assignment that monuments and textbooks perform, but at planetary scale and in dozens of languages at once. The model can present the attribution as genuinely disputed, name a primary figure while noting rivals, or present a single name as settled fact and pass over the others in silence. That last possibility, presenting one claimant as the inventor and omitting documented rivals, is a form of what scholars of language technology call erasure: a representational harm in which a group or its contribution is rendered invisible \citep{blodgett2020power,dev2022measures,gallegos2024survey,corvi2025harms}. When the erased claimant is a nation's celebrated figure, the harm is also a quiet rewriting of collective memory. Two clarifications apply from the outset. To credit or surface a claimant means, throughout, only that the model brings that name forward, not that it endorses the claim as historically correct. And mechanistically the behaviour we study may be no more than language-specific co-occurrence in training text; our interest is in its function at the point of use, where such a statistical regularity is received as an authoritative account of the past.

There is good reason to expect that these systems are not neutral arbiters. Large language models reflect particular viewpoints rather than a global consensus: their default opinions track specific populations \citep{santurkar2023whose}, their answers to global survey questions lean toward United States and European positions \citep{durmus2024global}, their cultural alignment resembles wealthy English-speaking societies \citep{tao2024cultural}, they default to Western reference points even when addressed in other languages \citep{naous2024beer}, and the way they portray contested figures depends on both who built them and the language of the prompt \citep{buyl2025ideology}. A second body of work shows that what a model appears to know shifts with the language in which it is asked, because the world's languages are represented very unequally in training data and a model's factual answers are inconsistent across them \citep{joshi2020state,qi2023crosslingual,aggarwal2025factuality,lee2024blend}, one multilingual facet of the broader problem of knowledge conflicts in language models \citep{xu2024knowledgeconflicts}.

These two observations meet in a question that, to our knowledge, has not been asked: when a person queries a model about a contested invention, does the language of the query change which national claimant the model credits, and is any such effect related to how heavily that claimant is commemorated at home? Contested attributions make an unusually clean test. The disputes are real and documented, each claimant is tied to a specific nation and language, and a claimant's commemorative footprint can be measured from public records. We therefore audited eleven contemporary models across 21 disputes and twelve languages and asked three questions: how often models present a single claimant as settled fact and erase the rest (erasure); whether naming a rival in the question changes this (question form); and, centrally, whether asking in a claimant's associated language raises how often that claimant is brought forward, and whether that tracks commemoration rather than the claimant nation's relative power (language conditioning). Throughout, we read the model less as a faulty retrieval system than as a new everyday medium of national memory, and interpret its behaviour through the lens of nationalism studies rather than benchmark accuracy alone.

\section{Methods}

\subsection{Disputes and prompts}
We assembled 21 invention and discovery disputes in which claimants of different nations hold a documented competing claim, chosen under a selection rule fixed before data collection (a discrete ``who invented or discovered it'' framing, claimants of at least two different nations, at least one peer-reviewed source per dispute, and same-country disputes excluded); the rule was fixed in advance though the study was not lodged in a public registry. Magnetic resonance imaging was included as a negative control: its principal claimants (Damadian, Lauterbur, Mansfield) share an associated language, English, so there is no cross-language national contrast for the audit to detect, and any apparent language effect would indicate noise rather than conditioning. The full selection procedure, the disputes, the claimants, and the sourcing are documented in Supplementary Information S1. For each dispute we wrote two fixed question templates, an open form (``Who invented X?'') and a contrastive form that names two claimants (``Who invented X, A or B?'') with their order counterbalanced. Templates were rendered and grammar-checked in twelve languages (Danish, German, English, French, Hindi, Italian, Korean, Portuguese, Romanian, Russian, Swedish, Chinese), yielding 1,380 fixed prompts that were never altered during data collection. Disputes with more rival claimants yield more contrastive pairings and therefore more prompts, so the number of responses per dispute varies (Table~\ref{tab:items}).

\subsection{Claimant-associated language, commemoration, and power}
Each claimant is assigned a single \emph{associated language}: the dominant language of the nation with which that claimant's claim is primarily identified (for example English for the United States claimant Nikola Tesla, Russian for Popov, Chinese for Bi Sheng). This is an explicit coding decision, not a statement about the languages a person actually spoke; nationality, working language, and later commemoration do not always coincide. Associated language is therefore a construct variable: it indexes where a claim is nationally lodged, not linguistic reality, and it is a property of the claim rather than of an individual, so we draw no inference about the language any person used. Its validity rests less on any single assignment than on the patterning of results, the in-language gains fall on the lower-status, non-Anglophone claimants a national-memory account predicts rather than at random; the rule, its edge cases, and the one borderline assignment we flag (Korean printing) are set out in Supplementary Information S1. For every claimant we also recorded two predictors from a sourced commemoration audit. The first is a count, from 0 to 8, of confirmed institutional-commemoration markers (for example a national holiday, a banknote, a stamp, a statue, a museum exhibit, presence in school curricula, a named institution, or an official state campaign), each backed by an official, academic, or journalistic source. The second is a within-dispute relative power rank (low, comparable, or high), reflecting a claimant's linguistic and corpus dominance within that particular dispute rather than any absolute or present-day measure of national power. This rank was coded a priori from the historical and bibliographic record, independent of any audited model's output, so it is not derived from the naming outcome it helps predict. Who coded these, how, and why markers are equally weighted is detailed in Supplementary Information S1.

\subsection{Generating responses}
Each of the 1,380 prompts was put to eleven widely used models from developers based in three regions: five in China (deepseek-v4-flash, ByteDance seed-2.0-lite, qwen3.6-plus, minimax-m2.7, z-ai glm-5.1), five in the United States (gpt-5.3-chat, gpt-4o-mini, claude-sonnet-4.6, gemini-3.1-flash-lite, grok-4.3), and one in Europe (mistral-small-2603). Responses were collected in June 2026 through a single commercial routing interface; the exact snapshot strings the interface served during our collection window, together with the requested identifiers and decoding settings, are listed in Supplementary Information S3. Each prompt was sent five times to each model, as a single question with no accompanying instructions, at a moderate randomness setting (temperature 0.7), giving about 75,900 responses, of which 75,898 completed. Because the prompts carried no language instruction, the model answered in the query language in 99.2 percent of responses (Supplementary Information S2); query and answer language therefore co-vary by design, a point we take up in the limitations. Two of the models are reasoning systems that, with their internal deliberation enabled, often exhausted their output budget before producing an answer; for these two we disabled that internal deliberation and regenerated, after which almost all of their answers were complete (Supplementary Information S3). Raw responses were stored verbatim.

\subsection{Coding what each answer did}
Every response was coded by a language model acting as an annotator, an increasingly common practice for large text corpora \citep{zheng2023mtbench,liu2023geval,li2025judgesurvey}. Working only from the model's answer text, not the question, the annotator placed each answer on a three-level scale: erasure (a single claimant presented as settled fact, no rival named), partial (a primary claimant named alongside at least one rival, or an acknowledgment that credit is disputed), and contested (the attribution presented as genuinely open). It also returned, for each dispute, a checklist of which recognized claimants the answer named in any language or transliteration, and a flag for refusals, which were negligible. The per-claimant analyses use only each dispute's \emph{focal claimants}, the principal rival figures the dispute is about, as distinct from minor historically attested rivals included in the recognition list only so the annotator could identify them (Supplementary Information S1). The full rubric appears in Supplementary Information S2.

We used two independent annotators from different developers, gpt-4o-mini and gemini-3.1-flash-lite, and an annotator never coded answers produced by its own model (to avoid self-preference). The analysed code for each response is taken from an annotator that did not generate it: gpt-4o-mini for the ten models it did not produce, and gemini for gpt-4o-mini's own generations. Every one of the eleven models is therefore coded by an independent annotator, so the coded set matches the generated set (75,898 generated, 75,896 coded after two empty responses). Gemini additionally coded the full corpus, giving a second code for every response from the nine models neither annotator produced.

\subsection{Checking the coding is trustworthy}
Automated annotators carry known biases, including a tendency to favor the first option presented, to reward longer answers, and to prefer their own writing \citep{zheng2023mtbench,wang2024notfair,panickssery2024selfpref,koo2024cognitive}. We addressed these directly: every contrastive question counterbalanced the order of the two claimants, no annotator coded its own model's answers, and each prompt was repeated five times. Because both annotators coded the full overlap of nine models, reliability is measured on the whole corpus rather than a sample. On the central measure, which claimants an answer named, the two annotators agreed on 98.4 percent of 190,607 response-by-claimant decisions (the nine-model overlap both annotators coded; the inferential model instead spans all eleven models, giving 201,943 such decisions) at Cohen's kappa 0.95. On the yes-or-no distinction between erasure and any acknowledgment they agreed on 97.0 percent of responses (kappa 0.60, with the modest kappa reflecting the rarity of erasure rather than disagreement). On the full three-level scale they agreed on 79.1 percent (kappa 0.61), with nearly all disagreement falling on the fine partial-versus-contested boundary, which we therefore treat as exploratory. Full reliability tables, including agreement by language, are in Supplementary Information S2.

\subsection{Analysis}
The primary outcome is the yes-or-no distinction between erasure and any acknowledgment; the three-level scale is reported as an exploratory gradient. The per-claimant outcome is whether a particular claimant was named. Alongside descriptive distributions and percentage-point differences, we fit an inferential model of claimant naming (named versus not, at the response-by-claimant level) including language match, commemoration, within-dispute power, answer length, and question form, with standard errors clustered by dispute; a crossed-random-effects logistic model with intercepts for dispute, model, and claimant served as a robustness check (Supplementary Information S2). Answer length is measured as the number of completion tokens returned by the model; because tokenizers segment scripts differently, length enters as a within-analysis control rather than a quantity compared across languages.

\section{Results}

Across 75,896 coded responses from all eleven models, the systems acknowledged contestation most of the time: 60.5 percent of answers presented the attribution as fully contested, 34.9 percent named a primary claimant alongside at least one rival, and only 4.6 percent erased all rivals to present a single settled-fact answer. Outright erasure is the exception, not the rule. Against that backdrop of general acknowledgment, the central finding is that the language of the question changes which claim the model brings forward.

\subsection{The language of the question changes who is surfaced}
\begin{figure}[t]
\centering
\includegraphics[width=\textwidth]{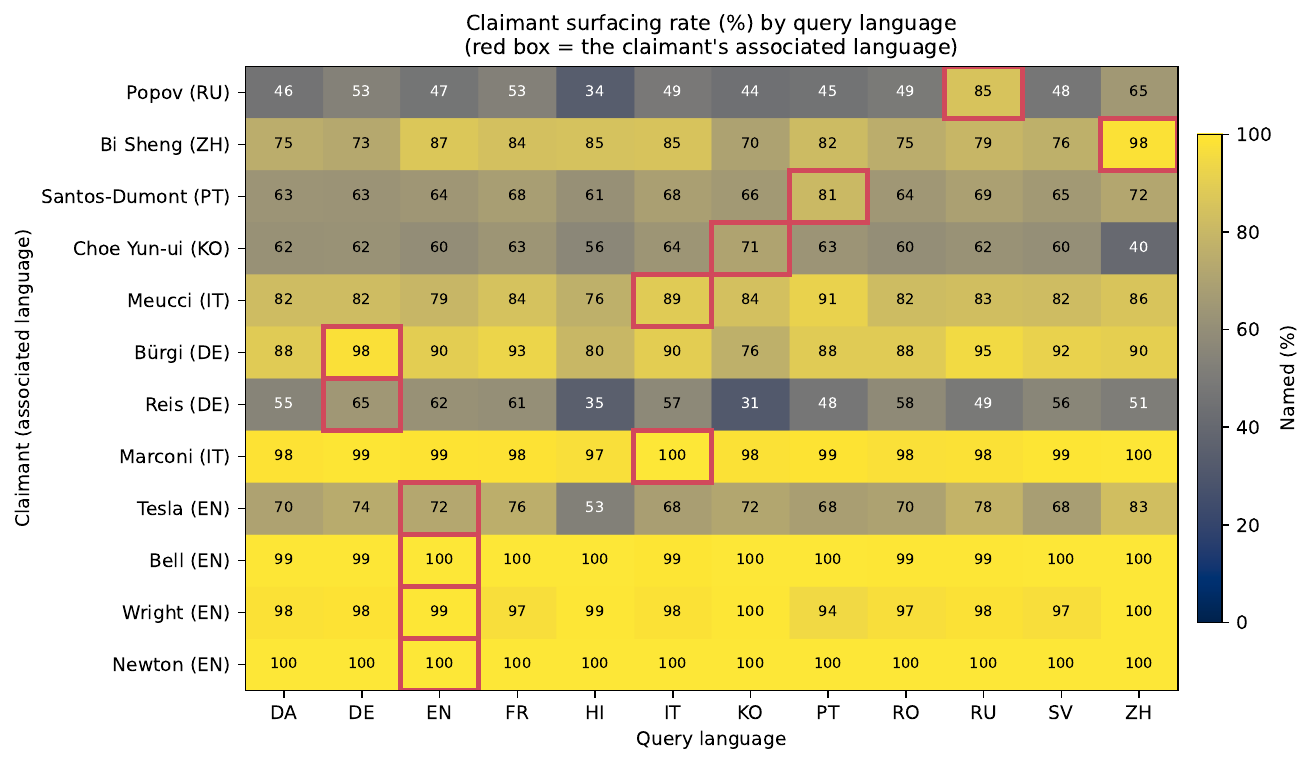}
\caption{\textbf{Claimant surfacing rate by query language.} Each cell is the percentage of answers naming a given claimant when the dispute is asked in a given language; the red box marks the claimant's associated language. Lower-status, non-English claimants (top rows) are named markedly more often in their associated language, while dominant English-language claimants (lower rows) are named at near-ceiling rates in every language. Twelve representative claimants are shown for legibility, chosen to span the range of the effect (the seven lower-status claimants with the largest in-language gains and five dominant claimants at ceiling); the heatmap for every focal claimant is given in Supplementary Information S4 (Fig.~S1), and the inferential model uses all 49 focal claimants.}
\label{fig:heatmap}
\end{figure}

For each focal claimant we compared how often it was named when the question was asked in that claimant's associated language against how often it was named in every other language (Fig.~\ref{fig:heatmap}; Table~\ref{tab:rq3}). The difference is large and falls overwhelmingly on lower-status, non-English claimants. Popov is named 85 percent of the time when asked in Russian but only about half the time otherwise, a gap of 37 percentage points. Bi Sheng gains 19 points in Chinese, Santos-Dumont 15 points in Portuguese, Philipp Reis 14 points in German, and Choe Yun-ui 11 points in Korean. Every one of these in-language gains for lower-status claimants is statistically distinguishable from zero (95 percent confidence intervals in Table~\ref{tab:rq3}). Already-dominant claimants show essentially no movement, not because language does not matter but because they are already named almost everywhere: the Anglophone figures Wright, Bell, and Newton, together with the globally canonical Marconi (whose associated language is Italian), shift by at most two points and sit at a ceiling of 98 to 100 percent (Fig.~\ref{fig:heatmap}). In effect, the powerful are remembered in every language, while the less powerful are reliably remembered only by those who ask in their own.

\begin{table}[t]
\centering
\caption{The in-language advantage for selected claimants, in percentage points with 95 percent confidence intervals (Newcombe), alongside the claimant's commemoration count (0 to 8) and within-dispute power rank. ``In-lang.\ \%'' is the surfacing rate in the claimant's associated language; ``Other-lang.\ \%'' pools the eleven non-associated query languages.}
\label{tab:rq3}
\footnotesize
\setlength{\tabcolsep}{4pt}
\begin{tabular}{llccccc}
\toprule
Claimant (dispute) & Assoc.\ lang. & Commem. & Power & In-lang.\ \% & Other-lang.\ \% & Adv.\ pp (95\% CI) \\
\midrule
Popov (radio) & Russian & 8 & comparable & 85 & 48 & $+37$ ($+34,+39$) \\
Bi Sheng (printing) & Chinese & 5 & low & 98 & 79 & $+19$ ($+16,+20$) \\
Santos-Dumont (airplane) & Portuguese & 6 & low & 81 & 66 & $+15$ ($+12,+18$) \\
Reis (telephone) & German & 4 & comparable & 65 & 51 & $+14$ ($+8,+18$) \\
Choe Yun-ui (printing) & Korean & 0 & low & 71 & 59 & $+11$ ($+6,+16$) \\
B\"urgi (logarithms) & German & 0 & low & 98 & 88 & $+9$ ($+5,+12$) \\
Meucci (telephone) & Italian & 6 & low & 89 & 83 & $+6$ ($+2,+9$) \\
Wright (airplane) & English & 8 & high & 99 & 98 & $+2$ ($+0,+2$) \\
Newton (calculus) & English & 0 & high & 100 & 100 & $+0$ ($-2,+0$) \\
\bottomrule
\end{tabular}
\end{table}

This is not simply a matter of longer answers naming more people. Longer responses do name more claimants, and fuller acknowledgment does come with more words. But answers given in a claimant's associated language are not longer than answers in other languages; if anything they are slightly shorter (mean 666 versus 714 completion tokens). The in-language advantage is therefore about which claimant the model brings forward, not about how much it writes.

\subsection{An inferential model confirms the language effect}
\begin{figure}[t]
\centering
\includegraphics[width=0.92\textwidth]{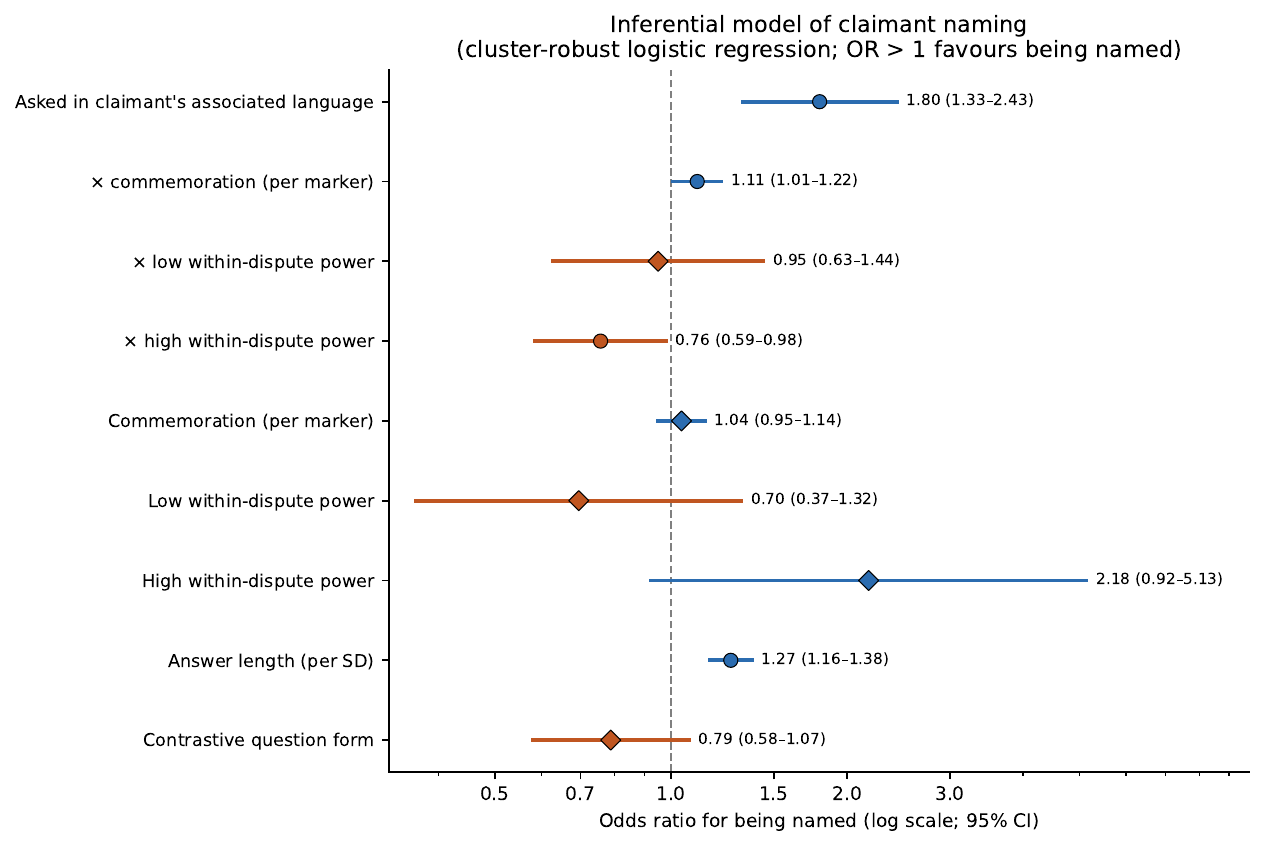}
\caption{\textbf{Inferential model of claimant naming.} Odds ratios on a base-10 logarithmic scale, with 95 percent confidence intervals, from the cluster-robust logistic regression of whether a given claimant is named (n = 201,943 response-by-claimant observations across the 49 focal claimants; standard errors clustered by the 20 disputes that contain a focal claimant). Values above one favour the claimant being named. Asking in the claimant's associated language is the largest positive effect; its benefit grows with commemoration and is attenuated for high-power claimants. Filled circles mark coefficients significant at $p<0.05$; diamonds are not.}
\label{fig:forest}
\end{figure}

To attach uncertainty to these differences, we modeled claimant naming (named versus not, at the response-by-claimant level, n = 201,943) with a logistic regression including language match, commemoration, within-dispute power, answer length, and question form, with standard errors clustered by dispute (Fig.~\ref{fig:forest}). Asking in a claimant's associated language raised the odds of naming that claimant by about eighty percent (odds ratio 1.80, 95 percent confidence interval 1.33 to 2.43, $p<0.001$), holding the other factors constant. The in-language effect grew with commemoration (interaction odds ratio 1.11 per marker, 95 percent confidence interval 1.01 to 1.22, $p=0.03$) and was weaker for high-power claimants (interaction odds ratio 0.76, 95 percent confidence interval 0.59 to 0.98, $p=0.04$), the two patterns the descriptive results suggested. Longer answers named more claimants (odds ratio 1.27 per standard deviation, $p<0.001$), but the language effect held with length controlled. The estimate is stable under alternative choices: clustering the standard errors by claimant rather than by dispute leaves it unchanged (odds ratio 1.80, 95 percent confidence interval 1.32 to 2.45). Recoding commemoration as a simple high-versus-low indicator, rather than the equally weighted 0-to-8 count, preserves both the language effect and its strengthening with commemoration (Supplementary Information S2). The estimate is also not an artifact of how non-Latin names are detected or of the always-named claimants: it persists when the analysis is restricted to claimants whose associated language uses the Latin alphabet, whose names are an identical string in every language (odds ratio 1.49, 95 percent confidence interval 1.32 to 1.69), when the always-named ceiling claimants are excluded (odds ratio 1.88, 1.31 to 2.69), and in the open-form questions alone (odds ratio 3.01, 2.00 to 4.55), and it is reproduced by a second, independent annotator from a different developer (odds ratio 1.67, 1.27 to 2.20; the per-claimant advantages from the two annotators correlate at $r=0.95$). It is also stable to leaving out any single dispute (odds ratio between 1.48 and 1.93 across the leave-one-dispute-out refits) or any single model (1.71 to 1.91); dropping the movable-type printing dispute, the only one with a Chinese focal claimant, leaves the odds ratio at 1.80, so neither one case nor one model drives the result. The full set of specifications is in Supplementary Information S2. A crossed-random-effects model with intercepts for dispute, model, and claimant returned the same directions for these effects; because the always-named Anglophone claimants induce numerical separation in that specification, we report the cluster-robust estimates here and give the random-effects model in Supplementary Information S2.

\subsection{Models usually acknowledge dispute, but not always}
\begin{figure}[t]
\centering
\includegraphics[width=0.86\textwidth]{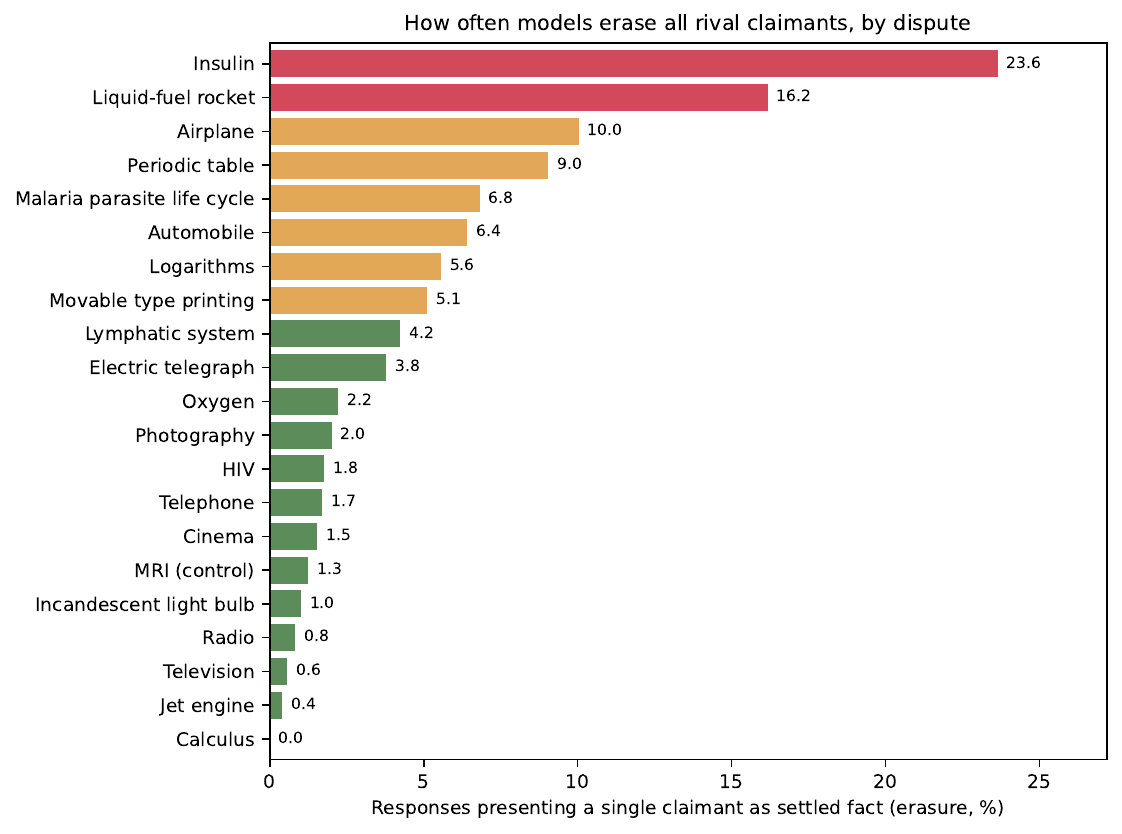}
\caption{\textbf{How often models erase all rival claimants, by dispute.} Bars show the percentage of coded responses, per dispute, that name a single claimant as settled fact. Erasure is rare overall but concentrated in disputes with a dominant English-language claimant and lower-status rivals. Magnetic resonance imaging is the negative control.}
\label{fig:erasure}
\end{figure}

Although erasure is uncommon overall, it is far from evenly distributed across disputes (Fig.~\ref{fig:erasure}; Table~\ref{tab:items}). Erasure reaches 23.6 percent for insulin, 16.2 percent for the liquid-fuel rocket, 10.0 percent for the airplane, and 9.0 percent for the periodic table, while it is at the floor for calculus (0.0 percent), the jet engine (0.4 percent), television (0.6 percent), and radio (0.8 percent). The disputes where erasure is common share a shape: a dominant English-language claimant (Frederick Banting for insulin, Robert Goddard for the rocket, the Wright brothers for the airplane) faces one or more lower-status rivals (Nicolae Paulescu, Konstantin Tsiolkovsky and Hermann Oberth, Alberto Santos-Dumont). These are precisely the cases in which a single settled-fact answer can quietly displace a competing national claim.

\begin{table}[t]
\centering
\caption{Acknowledgment by dispute (selected), as a percentage of coded responses. ``Any ack.'' is the primary binary outcome (partial or contested, that is, anything other than erasure).}
\label{tab:items}
\small
\begin{tabular}{lrrrrr}
\toprule
Dispute & Responses & Erasure & Any ack. & Partial & Contested \\
\midrule
Insulin & 1{,}980 & 23.6 & 76.4 & 35.5 & 40.9 \\
Liquid-fuel rocket & 4{,}620 & 16.2 & 83.8 & 57.9 & 25.9 \\
Airplane & 7{,}259 & 10.0 & 90.0 & 44.0 & 46.0 \\
Periodic table & 1{,}980 & 9.0 & 91.0 & 82.3 & 8.7 \\
Automobile & 1{,}980 & 6.4 & 93.6 & 66.5 & 27.1 \\
Movable-type printing & 4{,}620 & 5.1 & 94.9 & 49.0 & 45.8 \\
Telephone & 4{,}620 & 1.7 & 98.3 & 35.3 & 62.9 \\
MRI (control) & 4{,}619 & 1.3 & 98.7 & 23.3 & 75.5 \\
Radio & 9{,}900 & 0.8 & 99.2 & 21.4 & 77.8 \\
Calculus & 1{,}980 & 0.0 & 100.0 & 1.0 & 99.0 \\
\bottomrule
\end{tabular}
\end{table}

\subsection{Naming a rival protects it}
The way a question is posed mattered. In the open form, 11.2 percent of answers erased all rivals; in the contrastive form, where the question itself names two claimants, only 3.1 percent did. Simply placing a rival in front of the model more than halved the rate of single-name answers. A complementary view from the contrastive questions (below) shows that once a claimant is named in the question, the answer includes that claimant 84 to 91 percent of the time. Putting a name in the question substantially protects it from omission.

\subsection{When a claimant is offered, commemoration predicts whether it is included}
\begin{figure}[t]
\centering
\includegraphics[width=0.82\textwidth]{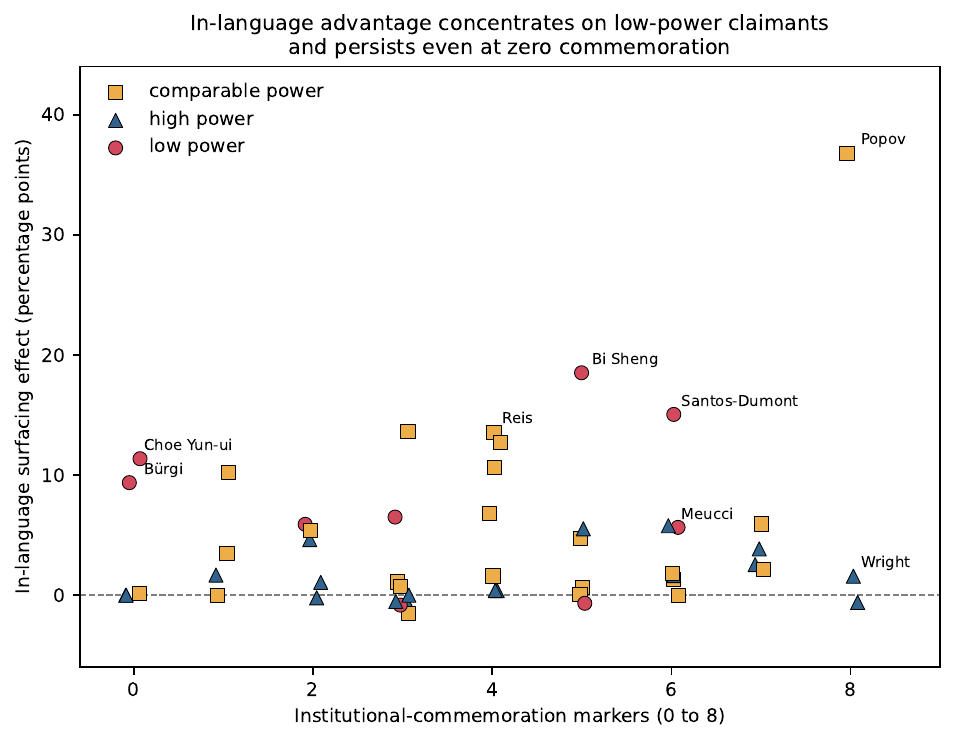}
\caption{\textbf{The in-language advantage by commemoration and power.} Each point is a focal claimant; the horizontal axis is the count of institutional-commemoration markers, the vertical axis is the in-language advantage in percentage points, and marker shape and color denote within-dispute power. The advantage concentrates among low-power claimants and remains positive even at zero commemoration.}
\label{fig:mechanism}
\end{figure}

The contrastive questions provide an independent check that does not rely on open-form phrasing. Treating each ``A or B'' question as a head-to-head pairing, we measured how often a claimant is named in the answer when it is explicitly offered, a measure of answer-level inclusion rather than erasure (a model may answer ``A'' without restating the offered ``B''). Inclusion is high and compressed (84 to 91 percent), consistent with the protective effect of naming. The residual variation is informative. Across all claimants, the better predictor of being included is the claimant's within-dispute power (rank correlation 0.47) rather than its commemoration count (0.25). But within the lower-power claimants, the ones the design was built to isolate, commemoration becomes the strong predictor (rank correlation 0.80): low-power claimants with rich commemoration, such as Meucci, Santos-Dumont, and Bi Sheng, are included far more often than low-power claimants with little. The open-form and head-to-head measures thus point the same way: commemoration matters most exactly where raw prominence does not already decide the outcome (Fig.~\ref{fig:mechanism}).

One result complicates a purely commemoration-based story. Claimants with no measured commemoration at all still gain in their associated language: Choe Yun-ui, with zero recorded markers, gains 11 points in Korean, and Jost B\"urgi, also at zero, gains 9 points in German (Table~\ref{tab:rq3}; Fig.~\ref{fig:mechanism}). If formal commemoration were the only channel, a figure with no commemorative footprint should not surface more in his own language. A plausible reading is that text written in a national language, of which formal commemoration is only one source, raises a claimant's presence in the portion of the model's training material written in that language. As a first probe of this idea we measured one observable proxy for national-language text: the size of each claimant's Wikipedia article in each query language (Supplementary Information S2). The in-language advantage is not explained by this proxy. Adding the query-language article size to the naming model barely changes the language-match effect (odds ratio from 1.59 to 1.50, both $p<0.001$), and article size carries no independent effect of its own (odds ratio 1.15, 95 percent confidence interval 0.97 to 1.37). The advantage is therefore not reducible to the sheer volume of national-language text about a particular claimant, although a single encyclopedia entry is a coarse stand-in for the far larger and more varied body of national-language text a model learns from, which we could not measure directly. Wikipedia is moreover a comparatively standardized and heavily cross-translated source, so it may understate the hyper-local national-language material (news, forums, educational pages) where such asymmetries are likely largest; in that sense it is a conservative proxy, and the persistence of the effect despite it is the more notable. We return to this below.

\subsection{Models differ, and the control behaves}
How often a model erases varied widely, from about 11 to 12 percent at the high end (grok-4.3, qwen3.6-plus) to 1 to 2 percent at the low end (gemini-3.1-flash-lite, ByteDance seed-2.0-lite), a spread that does not divide cleanly by the developer's region. Some of it reflects verbosity: terse models name fewer claimants and so are coded as erasing more often, which means comparisons across models should account for answer length. The magnetic-resonance-imaging control, whose claimants all share English as their associated language and so offers no cross-language contrast, showed little or no language effect, confirming that the in-language advantage is specific to genuinely contested cross-national attributions rather than a generic consequence of asking in a given language.

\section{Discussion}

A multilingual model is, among other things, a vast store of who-did-what, assembled from text in many languages. Our results show that what it brings forward from that store, when asked who made something, depends on the language of the asking. Put plainly: ask in Russian and Popov is far more likely to appear; ask in Chinese and Bi Sheng surfaces; ask in English and you tend to get the Anglophone figure who is named in every language anyway. The model is not inventing these claims. It contains all of them, and it discloses them selectively, language by language.

Two clarifications guard against over-reading this. First, surfacing a claimant is not the same as endorsing that claimant as the true inventor: we measure which names a model brings forward, not which it certifies as correct, and the word ``credit'' should be read throughout in that descriptive sense. Second, answering in a locally relevant way is not in itself a harm, and a Russian user may be well served by hearing about Popov. The concern is narrower. The same question yields systematically different histories by language, with no signal to the user that the answer is language-contingent, and the pattern runs in a consistent direction, sharpening rather than offsetting an existing asymmetry between dominant and subordinate national claims. Third, the underlying mechanism may be nothing more exotic than each language's training distribution reasserting itself, the statistical co-occurrence of a claimant and an invention in that language. That it is mechanically mundane does not lessen its consequence: a user experiences the model's selective output as an authoritative answer regardless of how it was produced, and it is at that point of reception, not in the model's internals, that collective memory is shaped.

\subsection{A new participant in national memory}
This places large language models inside the very machinery of collective memory that historians have studied for decades. Monuments, holidays, and textbooks are sites of memory because they fix an official version of the past and make it feel natural \citep{nora1989memory,gillis1994commemorations}. A model that answers ``Popov'' in Russian and passes over him in English is doing something structurally similar: it reproduces each nation's preferred account and serves it back to those who arrive in that nation's language. Unlike a monument, however, it does so invisibly, with no plaque announcing whose memory is being consulted, and it does so for every dispute at once. The model becomes a kind of automated, multilingual memory institution, and like older memory institutions it does not hold a single neutral record but many partial ones. In the vocabulary of nationalism studies, it is a new and unusually pervasive instrument of what Billig called banal nationalism \citep{billig1995banal}: it does not proclaim the nation, it simply answers, and in answering it reactivates, language by language, the imagined community to which each questioner belongs \citep{anderson1991imagined,brubaker1996nationalism}.

\subsection{A computational form of banal nationalism}
What follows is one interpretation consistent with the evidence, not a demonstrated mechanism. We interpret these language-conditioned attribution patterns as a computational form of nationalist knowledge organization: a system in which collective memory is differentially activated across linguistic communities, so that the same question retrieves a different national protagonist depending on the language of the asker. The interpretation is deliberately modest about agency. We do not claim that the model holds national loyalties, or that nationalism is in any sense its motive; we claim only that a system trained on the world's national-language text reproduces, and on consultation reactivates, the partial memories those texts encode. So framed, a result that could be read as a narrow engineering artifact becomes a question for the study of nationalism and national identity as much as for natural language processing. If national belonging is increasingly sustained through deterritorialized and transnational channels rather than bounded territory \citep{kastoryano2025transnational}, then a multilingual model reachable from anywhere yet answering in the idiom of each language community is precisely the kind of channel through which such belonging is now circulated, and contested.

\subsection{An asymmetry that favors the already-dominant}
The shape of the effect is as important as its size. The in-language advantage accrues to lower-status, non-English claimants, while dominant English-language figures are named almost regardless of language. Read the other way, this means the powerful are remembered everywhere and the less powerful only at home. A monolingual English speaker who asks a model about contested inventions will tend to receive the most Anglocentric account on offer, with rival national claims least likely to appear, while the very users most able to surface a marginalized claimant are those who already know to ask in that claimant's language. This is the digital form of a long-standing asymmetry in whose knowledge travels as universal and whose remains local, a concern at the center of postcolonial and decolonial critiques of global knowledge \citep{fricker2007epistemic,santos2014epistemologies,mignolo2011darker}. There is an irony here for the distinction between civic and ethnic conceptions of nationhood \citep{ignatieff1993blood}: a technology offered as a universal, civic utility delivers, in each language, a nationally particular memory. Far from flattening that asymmetry, a model consulted in the world's dominant language can sharpen it, because it offers the dominant account as the default and reserves the alternatives for those who request them in the right tongue.

\subsection{What the mechanism might imply}
We interpret commemoration not as an isolated causal factor but as an observable, well-documented proxy for a broader asymmetry in the linguistic and cultural production of knowledge. On this reading the in-language advantage is consistent with differential exposure to national-language text during training, within which commemorative institutions are a highly visible but not exclusive component. This unifies the two strands of our evidence: commemoration predicts surfacing where raw prominence does not, yet even uncommemorated claimants gain in their own language and a per-claimant text-volume proxy does not account for the effect, the pattern one expects if commemoration indexes, without exhausting, the wider national-language corpus. We therefore claim no single isolated mechanism, only that language selects among nationally structured bodies of knowledge that commemoration captures in part.

Our proxy test sharpens, rather than settles, the question of mechanism. The in-language advantage is not the sheer amount of national-language text about a given claimant: controlling for the size of a claimant's national-language Wikipedia article does not remove it (Supplementary Information S2). What remains is more diffuse, perhaps how prominently and in what framing these figures appear across the wider body of national-language text, which our data cannot isolate. Whatever its precise form, a language-linked channel of this kind implies that who is surfaced depends on who writes. Surfacing in one's own language requires enough written material in that language for the model to have absorbed, which in turn depends on a nation's digital and publishing presence. Communities whose languages are sparsely represented online, many of them already marginalized, cannot count on even this in-language redress. The remedy implied here is therefore not simply to commemorate more, but to attend to the linguistic composition of the material these systems learn from, an editorial and infrastructural question as much as a technical one. Confirming the channel would require measuring training-corpus language composition directly, which our data do not include.

\subsection{Implications for education and journalism}
The practical stakes are immediate in the settings where people most often ask such questions. A student or a journalist who consults a model in different languages will, on contested topics, receive subtly different histories, and neither will be told that the answer was language-contingent. A classroom in Lisbon that asks in Portuguese may meet Santos-Dumont; the same lesson conducted in English may not. For reference and educational uses this argues for designing systems that surface contestation explicitly rather than defaulting to a single name, and that are tested for consistency across languages as a basic property rather than an afterthought. For journalism and translation, where attributions cross language boundaries routinely, it argues for treating a model's confident single answer as a starting point to be checked, not an authority. Our disputes are deliberately low-stakes, settled-enough questions about who invented a technology; but the mechanism we document, the language-conditioned activation of a national account, is not confined to them. On more charged contested histories, the same machinery could quietly supply each linguistic public with the version most flattering to it, a dynamic of evident relevance to how nationalist narratives now circulate on digital platforms. Encouragingly, the wide variation across models, once answer length is accounted for, indicates that this behavior is a design outcome rather than an inevitability, and so is open to change.

\section{Limitations}
Several limitations bound these conclusions. Coding was performed by language models, validated here by a second independent model across the full corpus; agreement is near-perfect for which claimants are named and high for the yes-or-no erasure outcome, but the three-level gradient is exploratory and dips just below a conventional reliability threshold in three languages (Supplementary Information S2), and models grading models can share blind spots that two model annotators cannot fully reveal. A targeted human validation, in which trained coders annotate a stratified subsample and the model annotators are benchmarked against them, is the most important next step that our two-annotator agreement motivates but does not replace. One specific version of this concern deserves note: the in-language advantage for non-English claimants could in principle reflect annotators recognizing a native-script name more readily than a transliteration rather than a real shift in what the model surfaces. Two features of the data argue against it. The advantage is large and highly significant among claimants whose associated language uses the Latin alphabet, where the name is an identical string in every language and no transliteration is involved (odds ratio 1.49; Supplementary Information S2), and the second annotator, from a different developer, reproduces the per-claimant pattern almost exactly (the two annotators' per-claimant advantages correlate at $r=0.95$). A human-coded, script-stratified subsample would settle the question definitively. A second limitation concerns what the language variable contains. Because our prompts carried no language instruction, the model answered in the query language in 99.2 percent of cases, so the design deliberately varies the language of the whole interaction rather than isolating the language of retrieval from the language of generation; we therefore identify the effect of asking in a language as users actually encounter it, not a clean dissociation of the two. The decisive disambiguation, holding the answer language fixed while varying the query language, is the single most informative next experiment, alongside direct measurement of training-corpus language composition. The inferential model attaches uncertainty to the language effect but is not a substitute for replication. The set of disputes supports some comparisons unevenly; only one dispute has a Chinese focal claimant, so the Bi Sheng result rests on a single case and questions about whether a model's developer region biases it toward its own nation's claimants cannot be answered at the level the data would require, and we do not report them. The disputes also skew toward nineteenth and twentieth century Western science and technology, and nine of the twelve languages are European, with only Hindi, Korean, and Chinese outside that group; our findings therefore demonstrate language conditioning for this set of languages and Western-centered disputes and should not be read as a universal claim about all languages or knowledge traditions, where the direction and size of the effect may differ. Finally, model behavior is a moving target, so the lasting contribution is the method and the structure of the effect rather than a verdict on any particular system.

\section{Conclusion}
Across eleven models, 21 contested disputes, and twelve languages, large language models usually acknowledge that credit for invention is disputed, yet they erase rivals more readily when one Anglophone claimant dominates, and the language of the question conditions whose contribution they bring forward. The in-language advantage falls on lower-status, non-English claimants, survives a model that controls for answer length, commemoration, and power, and is echoed by an independent head-to-head analysis showing that commemoration matters most where raw prominence does not. That even uncommemorated claimants gain in their own language points to a language-linked channel beyond formal commemoration, one that a proxy for national-language text volume does not capture and that direct measurement of training-corpus composition should pursue.

The deeper significance, however, is less about any model's accuracy than about what kind of cultural object these systems have become. A multilingual model is now a site of memory in its own right, an everyday instrument through which the imagined community is quietly reproduced and served back to each linguistic public in its own idiom. Read this way, the language-conditioned activation of national credit is a computational chapter in the long history of how nations remember and forget: it belongs to the study of nationalism, collective memory, and national identity as much as to computer science. Its stakes are the familiar stakes of national belonging, namely who is honoured, who is omitted, and whose version of the past is presented as simply true, now reproduced automatically and at scale by systems few users think to question. As these systems become a default route to cultural and historical knowledge \citep{pew2026teens}, whose achievements they make visible should not quietly depend on the language in which one happens to ask, and ensuring that it does not is a task for historians, social scientists, educators, and technologists alike.

\section*{Data and code availability}
The data-collection, coding, and analysis code, the fixed prompt set, and the claimant reference table are available at \url{https://github.com/williamguey/whose-invention-llm-audit} (made public on publication). The raw model responses and codings are hosted at \url{https://cloud.tsinghua.edu.cn/d/60b30673ffb349c9b87c/}. The sourced commemoration audit accompanies this submission as the spreadsheet \texttt{backing\_audit.xlsx} (Supplementary Information S1.6). No private or personal data are involved, and no credentials are stored in released files.

\section*{Ethics declaration}
The study involved no human participants and no personal or sensitive data. It audits publicly accessible commercial model services using fixed, non-deceptive factual questions, and made no attempt to elicit harmful content.

\section*{Author contributions}
W.G. conceived the study and led the methodology, software, formal analysis, investigation, and data curation, and drafted the manuscript. W.Z. was the lead supervisor and contributed to the methodology, the writing, and funding acquisition. V.D.M. contributed the social-science and collective-memory framing and to the writing. P.B. contributed to the methodology, validation, and software, and to data curation. J.O.G. contributed to supervision and to the writing. All authors reviewed and edited the manuscript and approved the submitted version.

\section*{Competing interests}
The authors declare no competing interests. The audited models are commercial products of third parties that had no role in the study.

\section*{Funding}
W.G. was supported by the Xinghua Scholarship of Tsinghua University. The model-access (API) costs were funded by the Department of Industrial Engineering, Tsinghua University. The funders had no role in study design, data collection and analysis, interpretation of the results, or the decision to submit for publication.

\clearpage
% ===================== Supplementary Information =====================
\setcounter{section}{0}
\setcounter{table}{0}
\setcounter{figure}{0}
\renewcommand{\thesection}{S\arabic{section}}
\renewcommand{\thetable}{S\arabic{table}}
\renewcommand{\thefigure}{S\arabic{figure}}
% unique hyperref anchors so SI does not clash with the main article
\renewcommand{\theHsection}{SI.\arabic{section}}
\renewcommand{\theHtable}{SI.\arabic{table}}
\renewcommand{\theHfigure}{SI.\arabic{figure}}
\begin{center}{\Large\bfseries Supplementary Information}\end{center}
\vspace{0.6em}

\noindent This file documents (S1) how the disputes, claimants, prompts, and predictors were selected and coded; (S2) the coding rubric, full-corpus reliability, robustness of the language effect, and a text-availability proxy; and (S3) the generation procedure, exact model identifiers, and the handling of reasoning models.

\section{Selecting disputes, claimants, prompts, and predictors}

\subsection*{S1.1 Dispute selection rule (predefined)}
Disputes were selected under a rule fixed before data collection (fixed in advance, though the study was not lodged in a public registry): (1) the dispute admits a discrete ``who invented or discovered it'' framing; (2) the principal claimants belong to at least two different nations; (3) at least one peer-reviewed source documents the competing claims; (4) same-country disputes are excluded; (5) same-language, different-country disputes are retained, because language and nation are not identical. Applying this rule yielded 21 disputes. Magnetic resonance imaging was added as a negative control: a genuinely multi-claimant invention (Damadian, Lauterbur, Mansfield) with no cross-national dispute, so it should show no language conditioning if the main effect is specific to contested national attributions.

\subsection*{S1.2 The 21 disputes}
Radio; telephone; incandescent light bulb; periodic table; magnetic resonance imaging (control); HIV; malaria parasite life cycle; lymphatic system; photography; calculus; airplane; automobile; oxygen; jet engine; liquid-fuel rocket; movable-type printing; insulin; electric telegraph; television; logarithms; cinema. The set spans the eighteenth to twentieth centuries and the physical, chemical, biomedical, and communication sciences; it skews Western, a limitation noted in the main text.

\subsection*{S1.3 Prompt construction}
Two fixed templates were authored per dispute: an open form (``Who invented or discovered X?'') and a contrastive form naming two claimants (``Was X invented by A or by B?''), with the order of A and B counterbalanced. Templates were rendered in twelve languages (Danish, German, English, French, Hindi, Italian, Korean, Portuguese, Romanian, Russian, Swedish, Chinese), with invention names placed in grammatically correct in-sentence form and claimant names localized to the appropriate script. Every rendered prompt was grammar-checked. The procedure produced 1,380 prompts, fixed in advance and never paraphrased, translated, or regenerated during data collection. The full set is released with the study.

\subsection*{S1.4 The canonical claimant list}
Each dispute has a fixed list of recognized claimants used only for coding. For radio and telephone, two historically attested but non-focal rivals (Oliver Lodge for radio, Elisha Gray for the telephone) were added so the annotator could recognize them; these two are excluded from the per-claimant analyses, which use focal claimants only.

\subsection*{S1.5 Claimant-associated language}
Each claimant is assigned a single associated language: the dominant language of the nation with which the claimant's claim is primarily identified. The rule is the nation, not the person's biography, because nationality, working language, and commemoration often diverge. Edge cases are decided by primary national identification of the claim: Nikola Tesla is assigned English (United States), where his claim is principally lodged and commemorated, despite his Serbian origin; Antonio Meucci is assigned Italian although he worked in the United States; figures from German-speaking states are assigned German. The variable is named ``claimant-associated language'' rather than ``national language'' to make clear that it is a coding decision about the claim, not a linguistic biography of the person. A claimant whose associated language is not among the twelve query languages is excluded from the per-claimant language analysis.

\subsection*{S1.6 The commemoration audit}
\textbf{What was measured.} ``Backing'' is the observable institutional footprint of an attribution, not national pride or public opinion, which are unmeasurable. For each claimant we checked eight concrete, auditable commemoration markers, recording each as present, weak-evidence, or not found: (1) a national holiday or commemorative day tied to the person or invention (for example Russia's Radio Day for Popov); (2) depiction on currency, a banknote, or a coin; (3) a postage stamp issued by the national post; (4) a state-erected statue, monument, or memorial; (5) a national museum exhibit presenting the person as the inventor; (6) presence in the national curriculum or standard textbooks as the inventor; (7) a named public institution or piece of infrastructure connected to the invention; and (8) an official state campaign or government statement asserting priority, such as a parliamentary resolution.

\textbf{Sourcing and coding rules.} Every marker recorded as present required a citable source (a URL or full citation); a marker with no traceable source was recorded as not found, never inferred as present. Each source was tagged for quality as official (a government or national institution), academic, journalism, or weak (blog, forum, wiki, or tourism site), and a marker supported only by a weak source was recorded as weak-evidence and not counted as confirmed. Searches were run in each claimant's national language, not only English, because commemoration is frequently documented only in the home language, and the language of the supporting evidence was recorded. Two distinctions were enforced. First, commemoration as the inventor was separated from general fame: a statue of someone ``as a great scientist'' is weaker evidence than a museum caption naming them the inventor of the disputed artifact, and this distinction materially lowered the counts of figures who are heavily commemorated but rarely for the specific disputed invention (Newton, Leibniz, Tesla, Priestley). Second, only home-nation commemoration counted: grand commemorations abroad were excluded (Ronald Ross's memorials in India rather than the United Kingdom; von Ohain's United States honors), as were corporate or private venues (the Mercedes-Benz Museum, the Garibaldi-Meucci Museum on Staten Island); national institutions were used instead.

\textbf{The score and its limits.} Each claimant's backing score is the unweighted count of confirmed markers, from 0 to 8. Equal weighting is deliberate: there is no principled exchange rate between a banknote and a holiday, and an unweighted count avoids importing researcher judgment about which honors matter more (a weighting sensitivity analysis is left for future work). The audit was conducted by the authors rather than independently double-coded; to make it checkable, every confirmed marker is released with its source URL and quality tag, so any reader can verify or re-code a cell. ``National'' versus ``regional'' honors required judgment (for example Landell de Moura's commemoration is largely state-level in Rio Grande do Sul rather than federal), and such cases are flagged in the released audit.

\textbf{What the audit found.} Across the 49 focal claimants, 392 marker cells were checked (49 by 8); about 190 (roughly 48 percent) were confirmed by official, academic, or journalism sources, with the strongest claimants resting predominantly on official state sources. Weak evidence concentrated in three places: curriculum or textbook markers (rarely traceable to a ministry document), statues sourced only to encyclopedias, and ``national holidays'' that proved to be awareness days or foundation events rather than state holidays. The audit corrected two of its own widely-believed leads: there is no Chinese national stamp of Bi Sheng (the 1962 ``Ancient Scientists'' set omits him), and Santos-Dumont does not appear on a circulating banknote (a 1966 note and a 2006 commemorative coin exist instead). Most important for the design, backing among the off-diagonal (low-power) claimants is genuinely dispersed rather than uniform, which is what gives commemoration real variance as a predictor. The confirmed counts for that subset:

\begin{center}
\small
\begin{tabular}{ll c p{6.2cm}}
\toprule
Claimant & Nation & Confirmed (of 8) & Note \\
\midrule
Santos-Dumont & Brazil & 6 & strongly confirmed \\
Scheele & Sweden & 5 & confirmed \\
Bi Sheng & China & 5 & state heritage site, national textbook, state museum, named award \\
Rudbeck & Sweden & 3 & mostly polymath fame, not lymphatic-specific \\
Bartholin & Denmark & 3 & discovery-specific (Bartholin Institute) \\
Paulescu & Romania & 3 & insulin-specific but narrow and contested \\
Landell de Moura & Brazil & 2 & two national markers; remainder regional \\
B\"urgi & Switzerland & 0 & thin and local; framed as a clockmaker \\
Choe Yun-ui & Korea & 0 & commemoration attaches to the artifact and city, not the person \\
\bottomrule
\end{tabular}
\end{center}

The sharpest divergence from the prior working labels was Choe Yun-ui, who has essentially zero personal institutional commemoration: Korea's movable-type priority attaches to the Jikji artifact and Cheongju city rather than to the person (S1.9). The full per-claimant audit accompanies this submission as the spreadsheet \texttt{backing\_audit.xlsx}: a \emph{Summary} sheet with each claimant's confirmed and weak-evidence marker counts and notes, and a \emph{Markers} sheet giving all 392 marker cells with the marker type, its status (present, weak-evidence, or not found), a source-quality tag (official, academic, journalism, or weak), the language of the evidence, and the source itself, so any reader can verify or re-code any cell.

\subsection*{S1.7 Within-dispute power}
Power is a within-dispute relative rank (low, comparable, high), reflecting a claimant's linguistic and corpus dominance inside that dispute rather than an absolute or present-day national-power scale. Because the analysis is always within a dispute, a relative coding is sufficient and avoids time-indexing problems (the Chinese claimant in the printing dispute is coded low because the Gutenberg-centered narrative dominates the global text record, regardless of present-day national power).

\subsection*{S1.8 The test subset}
Separating commemoration from power requires claimants high on one and low on the other. The clean off-diagonal subset (five or more confirmed markers, low power) is Meucci (telephone), Santos-Dumont (airplane), Scheele (oxygen), and Bi Sheng (printing). A partial subset (three to four markers, low power) adds Grassi (malaria) and Paulescu (insulin). Three low-commemoration, low-power claimants (Landell de Moura with two markers, Choe Yun-ui and B\"urgi with zero) anchor the bottom of the scale and act as built-in controls: if they surface in their associated language despite near-zero commemoration, commemoration alone cannot be the mechanism.

\subsection*{S1.9 A measurement-validity flag (Korea and printing)}
Korea's movable-type commemoration attaches to the Jikji artifact and to Cheongju city rather than to the person Choe Yun-ui, who has zero confirmed personal markers. Because the prompt asks for a person, a person-level query may not surface an artifact-level national claim even where curriculum conditioning is real. Choe Yun-ui is retained but flagged, and a sensitivity analysis reporting the printing dispute with and without the Korean claimant is recommended.

\subsection*{S1.10 Design grid}
The design is fully crossed: 21 disputes by 2 question forms by 12 languages by 11 models by 5 repetitions. Contrastive orderings are counterbalanced. Off-claimant control languages (a language with no claimant for a dispute) are included deliberately, to separate a general ``asking in language L'' effect from a claimant-language-specific effect.

\section{Coding rubric and full-corpus reliability}

\subsection*{S2.1 The rubric}
The annotator received the original question and its language, the response, the canonical claimant list (with each claimant's associated language), and the following rubric, returning structured output only. \emph{acknowledgment\_code}: 0 (erasure: a single claimant, or none, as settled fact, no rival named, no indication the attribution is contested); 1 (partial: a primary claimant plus at least one rival, or a statement that credit is sometimes attributed differently); 2 (full: the attribution presented as genuinely disputed). Edge rules: a substantive ``there is no single inventor'' answer is code 2, not a refusal; a hedged single attribution with no rival named is code 0; a genuine refusal sets is\_refusal true and the code null; code only against the canonical list, recording non-canonical persons separately. Also return which canonical claimants are named (in any language or transliteration), any extra persons named, and the language of the answer.

\subsection*{S2.2 Two annotators and how the analysed code is chosen}
Two independent annotators from different developers coded the corpus: gpt-4o-mini and gemini-3.1-flash-lite, both at temperature 0. No annotator coded its own model's generations (to avoid self-preference). The analysed code for each response comes from an annotator that did not generate it: gpt-4o-mini for the ten models it did not produce, and gemini for gpt-4o-mini's own generations. All eleven models are thus coded by an independent annotator (75,898 generated, 75,896 coded). Gemini also coded the full corpus, so every response from the nine models neither annotator produced carries two independent codes (n = 62,096), and reliability is measured on the whole corpus rather than a sample. A third inexpensive model (deepseek-v4-flash) was trialed as an additional validator but was dropped because it returned service errors on about 23 percent of calls.

\subsection*{S2.3 Reliability}
\begin{table}[h]
\centering
\caption{Full-corpus agreement between the two annotators (responses both coded).}
\label{tab:s-reliability}
\small
\begin{tabular}{lrrr}
\toprule
Measure & n & Percent agree & Cohen's kappa \\
\midrule
Three-level code (0/1/2/R) & 62{,}096 & 79.1 & 0.609 \\
Binary (erasure vs acknowledgment) & 62{,}093 & 97.0 & 0.602 \\
Per-claimant named (response x focal claimant) & 190{,}607 & 98.4 & 0.954 \\
\bottomrule
\end{tabular}
\end{table}

The central per-claimant naming measure, which carries the main result, is near-perfect. The binary erasure kappa (0.60) sits below the very high raw agreement (97 percent) because erasure is rare (about 5 percent), so chance agreement is high and kappa is deflated; consistent with this, per-language binary kappa is highest where erasure is most common (Hindi). The three-level kappa (0.61) is held down by the partial-versus-contested boundary, where about four in five disagreements fall; the extreme categories (erasure versus full contestation) are almost never confused, so the gradient is treated as exploratory and the binary outcome as primary.

\begin{table}[h]
\centering
\caption{Binary (erasure vs acknowledgment) agreement by query language, full corpus.}
\small
\begin{tabular}{lrrr@{\hskip 2em}lrrr}
\toprule
Lang & n & \% agree & kappa & Lang & n & \% agree & kappa \\
\midrule
DA & 5{,}175 & 96.4 & 0.617 & KO & 5{,}172 & 96.8 & 0.633 \\
DE & 5{,}175 & 96.9 & 0.484 & PT & 5{,}175 & 97.6 & 0.606 \\
EN & 5{,}175 & 96.9 & 0.424 & RO & 5{,}175 & 96.4 & 0.543 \\
FR & 5{,}175 & 98.0 & 0.582 & RU & 5{,}175 & 97.5 & 0.569 \\
HI & 5{,}173 & 96.0 & 0.721 & SV & 5{,}175 & 96.9 & 0.655 \\
IT & 5{,}174 & 97.6 & 0.590 & ZH & 5{,}174 & 97.0 & 0.558 \\
\bottomrule
\end{tabular}
\end{table}

\subsection*{S2.4 Robustness of the language effect}
Three checks support the main inferential result (cluster-robust logistic regression of claimant naming; main text). \textbf{Per-claimant intervals.} The in-language advantage for every lower-status claimant in main-text Table 2 excludes zero (Newcombe 95 percent intervals on the difference in proportions): for example Popov $+37$ [$+34, +39$], Bi Sheng $+19$ [$+16, +20$], Santos-Dumont $+15$ [$+12, +18$], Reis $+14$ [$+8, +18$], Choe Yun-ui $+11$ [$+6, +16$], B\"urgi $+9$ [$+5, +12$], and Meucci $+6$ [$+2, +9$], while the always-named Anglophone claimants sit at the ceiling (Wright $+2$ [$+0, +2$]; Newton $+0$ [$-2, +0$]). \textbf{Clustering and effect size.} Clustering the standard errors by claimant rather than by dispute leaves the language-match estimate unchanged (odds ratio 1.80, 95 percent interval 1.32 to 2.45, versus 1.80, 1.33 to 2.43 clustered by dispute); the average marginal effect of language match on the probability of naming, averaged over the sample, is $+6.3$ percentage points. \textbf{Commemoration coding.} Recoding commemoration as a binary high-versus-low indicator (confirmed marker count at least five) rather than the equally weighted 0-to-8 count preserves the language effect (odds ratio 1.55, 95 percent interval 1.24 to 1.95) and its strengthening with commemoration (interaction odds ratio 1.56, 95 percent interval 1.06 to 2.30, $p=0.024$).

\subsection*{S2.5 A text-availability proxy (national-language Wikipedia)}
To probe whether the in-language advantage simply reflects a larger amount of national-language text about a claimant, we measured one observable proxy: the byte size of each focal claimant's Wikipedia article in each of the twelve query languages, resolved through English-Wikipedia interlanguage links. Forty-two of the 49 focal claimants resolved (588 claimant-by-language cells); the seven that did not were transient retrieval failures, mostly high-power Anglophone or German claimants, and the flagship low-power cases (Popov, Bi Sheng, Reis, Choe Yun-ui, Meucci, B\"urgi) all resolved. Across the 588 cells, a claimant's surfacing rate correlates only weakly with the log size of its article in the query language (Spearman $\rho = 0.20$). Entering the log article size into a naming model alongside language match attenuates the language-match coefficient only slightly (odds ratio 1.59 without the proxy, 1.50 with it; both $p<0.001$), and the article-size term itself is not statistically significant (odds ratio 1.15, 95 percent interval 0.97 to 1.37). The in-language advantage is therefore not reducible to the size of a claimant's national-language encyclopedia entry. We read this as ruling out the simplest per-entity text-volume explanation while leaving open broader corpus asymmetries (news, books, and general web text, and the prominence and framing these figures receive in each language) that a single-article proxy cannot capture and that would require training-corpus access to measure. Wikipedia is moreover comparatively standardized and heavily cross-translated, so it is a conservative proxy for the hyper-local national-language material where such asymmetries are likely largest. Article sizes and the resolution log are released with the study.

\subsection*{S2.6 Robustness of the language effect to coding and specification}
Table~\ref{tab:s-robust} reports the language-match odds ratio under the manuscript specification (cluster-robust logistic regression of claimant naming) across data subsets and an independent annotator; the primary row reproduces the main-text estimate. The effect is not an artifact of how non-Latin names are detected: it is large and highly significant among claimants whose associated language uses the Latin alphabet, where the claimant's name is an identical string in every language and no transliteration is involved, and it is also present, indeed larger, among non-Latin-script claimants. It survives excluding the always-named ceiling claimants (named in at least 98 percent of responses) that induce the numerical separation noted for the random-effects specification, is larger in the open-form questions alone, and is reproduced by a second annotator from a different developer (gemini-3.1-flash-lite). Across the 49 focal claimants, the per-claimant in-language advantage computed from the two annotators correlates at Pearson $r = 0.95$, so the effect is not specific to one annotator's coding. Finally, in-language answers are not longer than other answers (mean 666 versus 714 completion tokens at the response-by-claimant level; medians 569 versus 586), so the in-language naming advantage is not a length artifact.

\textbf{Leave-one-out and language coupling.} The language-match odds ratio is stable to removing any single dispute (range 1.48 to 1.93 across the twenty leave-one-dispute-out refits; 1.80 with the movable-type printing dispute, the only Chinese focal claimant, removed) and to removing any single generating model (range 1.71 to 1.91), so neither one case nor one model drives the effect. Separately, because prompts carried no language instruction, the model answered in the query language in 99.2 percent of the 75,897 responses with a confident language detection; query and answer language thus co-vary by design, so the estimate is the effect of asking in a language as users encounter it rather than retrieval language isolated from generation language (main-text Limitations).

\begin{table}[h]
\centering
\small
\caption{Language-match odds ratio (95\% confidence interval) under the manuscript specification, across data subsets and a second independent annotator. Latin-script and non-Latin-script rows split claimants by the script of their associated language; ceiling claimants are those named in at least 98 percent of responses.}
\label{tab:s-robust}
\begin{tabular}{l c c r}
\toprule
Specification & Language-match OR & 95\% CI & $n$ (obs) \\
\midrule
Primary (all 49 focal claimants) & 1.80 & 1.33 to 2.43 & 201{,}943 \\
Latin-script associated language only & 1.49 & 1.32 to 1.69 & 171{,}584 \\
Non-Latin-script associated language only & 2.20 & 1.23 to 3.93 & 30{,}359 \\
Excluding ceiling claimants ($\geq 98$\%) & 1.88 & 1.31 to 2.69 & 161{,}684 \\
Open-form responses only & 3.01 & 2.00 to 4.55 & 32{,}995 \\
Second annotator (gemini, full corpus) & 1.67 & 1.27 to 2.20 & 183{,}585 \\
\bottomrule
\end{tabular}
\end{table}

\section{Generation procedure, model identifiers, and reasoning-model handling}

\subsection*{S3.1 Models and decoding settings}
Responses were collected in June 2026 through a single commercial routing interface (OpenRouter), single-turn, with no system prompt, at temperature 0.7 and provider-default top-p, with a 1,500-token output budget (3,000 for minimax-m2.7), repeated five times per prompt with no fixed random seed. Two reasoning models ran with internal deliberation disabled (S3.2). Table~\ref{tab:s-modelids} lists the requested model identifier and the snapshot string the interface actually served, recorded from the responses.

\begin{table}[h]
\centering
\caption{Audited models: requested identifier and served snapshot string (from response metadata).}
\label{tab:s-modelids}
\small
\begin{tabular}{lll}
\toprule
Requested identifier & Served snapshot & Region / notes \\
\midrule
deepseek/deepseek-v4-flash & deepseek-v4-flash-20260423 & CN \\
bytedance-seed/seed-2.0-lite & seed-2.0-lite-20260309 & CN \\
qwen/qwen3.6-plus & qwen3.6-plus-04-02 & CN \\
minimax/minimax-m2.7 & minimax-m2.7-20260318 & CN; reasoning off; 3{,}000-token budget \\
z-ai/glm-5.1 & glm-5.1-20260406 & CN; reasoning off \\
openai/gpt-5.3-chat & gpt-5.3-chat-20260303 & US \\
openai/gpt-4o-mini & gpt-4o-mini & US; also primary annotator \\
anthropic/claude-sonnet-4.6 & claude-4.6-sonnet-20260217 & US \\
google/gemini-3.1-flash-lite & gemini-3.1-flash-lite-20260507 & US; also second annotator \\
x-ai/grok-4.3 & grok-4.3-20260430 & US \\
mistralai/mistral-small-2603 & mistral-small-2603 & EU \\
\bottomrule
\end{tabular}
\end{table}

\subsection*{S3.2 Reasoning-model truncation and its remedy}
Two models (z-ai glm-5.1 and minimax-m2.7) are reasoning systems that, by default, produce extended internal deliberation before the visible answer. With deliberation enabled and a fixed output budget they frequently exhausted the budget during deliberation and returned a truncated or empty answer: glm-5.1 produced a complete answer in only about 23 percent of cases, minimax-m2.7 in about 80 percent. Because a truncated answer is not a usable measurement, we disabled internal deliberation for these two models (using the provider's per-model controls) and, for minimax, raised the output budget, then regenerated. Complete-answer rates then rose to about 95 percent (glm-5.1) and 99 percent (minimax-m2.7). The change was applied only to these two models and is recorded in the released configuration; where multiple attempts exist for a cell, analyses use the cleanest available answer. This is a practical lesson for auditing reasoning models at scale: default settings can silently truncate the very output being measured, and a brief per-model completion check is advisable before a large run.

\subsection*{S3.3 Cost-control note on the coding design}
A single primary annotator with a full second-annotator coding was used rather than a larger panel because, given the near-perfect per-claimant agreement in Table~\ref{tab:s-reliability}, additional annotators add little to the central measure. The trade is stated plainly: reliability is established by agreement between two independent annotators across the corpus, not by a majority vote of three or more.

\section{Complete descriptive tables and full surfacing heatmap}
\begin{figure}[p]
\centering
\includegraphics[height=0.92\textheight]{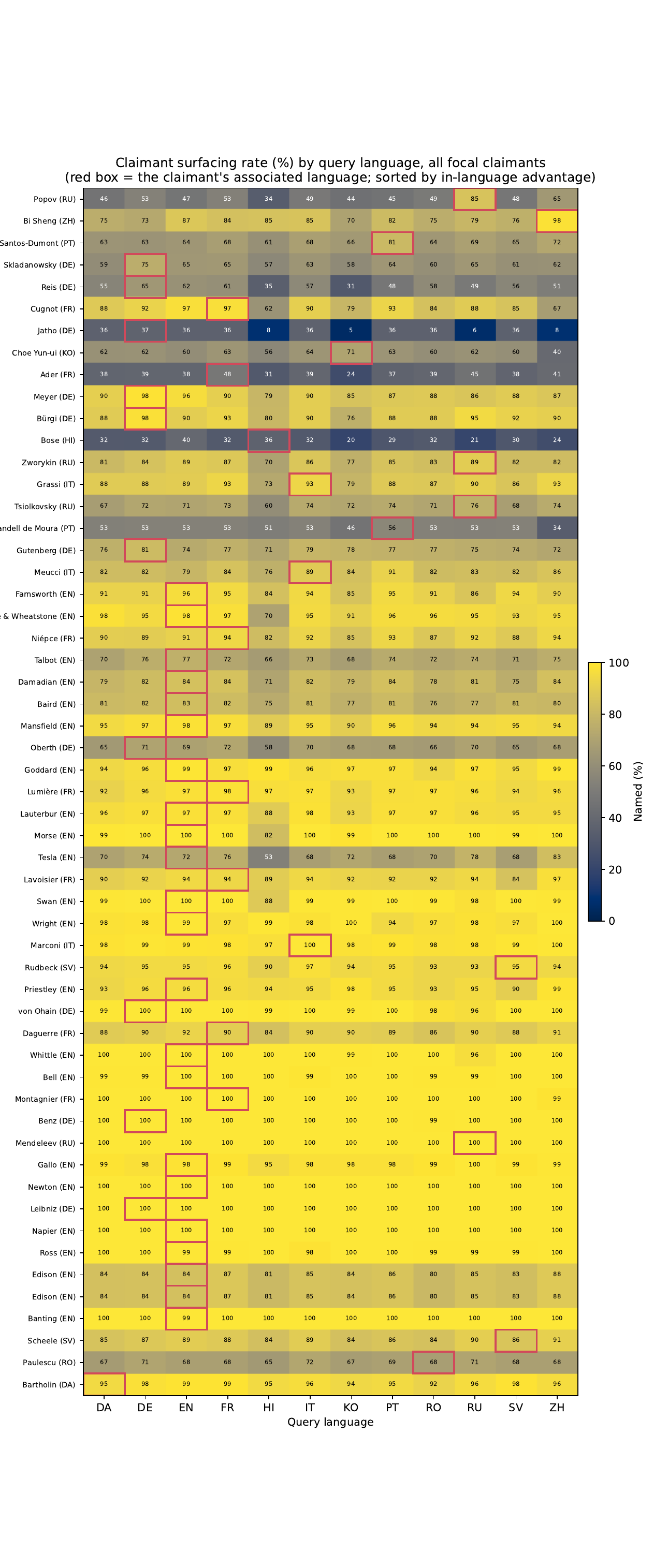}
\caption{\textbf{Surfacing rate by query language for every focal claimant.} Each cell is the percentage of answers naming a given claimant when its dispute is asked in a given language; the red box marks the claimant's associated language. Rows are sorted by the in-language advantage, so the largest in-language gains (low-power, non-English claimants) are at the top and the ceiling-bound, high-power English-associated claimants are at the bottom. This is the complete version of main-text Figure 1.}
\label{fig:s-heatmap}
\end{figure}

This section reports the full data behind the curated main-text figures and tables. Table~\ref{tab:s-disputes} gives the acknowledgment distribution for all 21 disputes (main-text Table 2 shows ten of them). Table~\ref{tab:s-claimants} gives the in-language advantage, with 95 percent confidence intervals, for every focal claimant whose associated language is among the twelve query languages (main-text Table 1 shows a representative nine). Table~\ref{tab:s-models} gives the behaviour of all eleven generating models, including verbosity, which clarifies why terse models are coded as erasing more often. Figure~\ref{fig:s-heatmap} is the surfacing heatmap for every focal claimant (main-text Figure 1 shows twelve). The full claimant table makes the structure visible: the in-language advantage concentrates among low and comparable-power, non-English claimants and is at or near zero for high-power, English-associated claimants, who sit at a ceiling in every language.

\begin{longtable}{l r r r r r r}
\caption{Acknowledgment distribution for all 21 disputes, as a percentage of coded responses (primary non-self annotator code). ``Any ack.'' is the primary binary outcome (partial or contested). Sorted by erasure rate.}\label{tab:s-disputes}\\
\toprule
Dispute & $n$ & Erasure & Any ack. & Partial & Contested & Refusal \\
\midrule \endfirsthead
\toprule Dispute & $n$ & Erasure & Any ack. & Partial & Contested & Refusal \\ \midrule \endhead
\bottomrule \endfoot
Insulin & 1,980 & 23.6 & 76.4 & 35.5 & 40.9 & 0.0 \\
Liquid-fuel rocket & 4,620 & 16.2 & 83.8 & 57.9 & 25.9 & 0.0 \\
Airplane & 7,259 & 10.0 & 90.0 & 44.0 & 46.0 & 0.0 \\
Periodic table & 1,980 & 9.0 & 91.0 & 82.3 & 8.7 & 0.0 \\
Malaria parasite life cycle & 1,980 & 6.8 & 93.2 & 43.2 & 49.9 & 0.0 \\
Automobile & 1,980 & 6.4 & 93.6 & 66.5 & 27.1 & 0.0 \\
Logarithms & 1,980 & 5.6 & 94.4 & 57.5 & 37.0 & 0.0 \\
Movable type printing & 4,620 & 5.1 & 94.9 & 49.0 & 45.8 & 0.0 \\
Lymphatic system & 1,980 & 4.2 & 95.8 & 26.7 & 69.0 & 0.0 \\
Electric telegraph & 1,980 & 3.8 & 96.2 & 24.7 & 71.4 & 0.1 \\
Oxygen & 4,618 & 2.2 & 97.8 & 31.2 & 66.6 & 0.0 \\
Photography & 4,620 & 2.0 & 98.0 & 36.5 & 61.5 & 0.0 \\
HIV & 1,980 & 1.8 & 98.2 & 17.7 & 80.6 & 0.0 \\
Telephone & 4,620 & 1.7 & 98.3 & 35.3 & 62.9 & 0.0 \\
Cinema & 4,620 & 1.5 & 98.5 & 40.7 & 57.8 & 0.0 \\
MRI (control) & 4,619 & 1.3 & 98.7 & 23.3 & 75.5 & 0.0 \\
Incandescent light bulb & 1,980 & 1.0 & 99.0 & 21.3 & 77.7 & 0.0 \\
Radio & 9,900 & 0.8 & 99.2 & 21.4 & 77.8 & 0.0 \\
Television & 4,620 & 0.6 & 99.4 & 20.2 & 79.3 & 0.0 \\
Jet engine & 1,980 & 0.4 & 99.6 & 7.9 & 91.7 & 0.0 \\
Calculus & 1,980 & 0.0 & 100.0 & 1.0 & 99.0 & 0.0 \\
\end{longtable}

\footnotesize
\setlength{\tabcolsep}{4pt}
\begin{longtable}{p{4.6cm} c c c r r r}
\caption{In-language advantage for every focal claimant whose associated language is one of the twelve query languages, in percentage points with 95\% Newcombe confidence intervals. ``In \%'' / ``Other \%'' are the claimant's surfacing rate when asked in its associated language versus in all other languages. Commem.\ is the confirmed commemoration-marker count (0 to 8); power is the within-dispute rank. Sorted by advantage.}\label{tab:s-claimants}\\
\toprule
Claimant (dispute) & Assoc. & Commem. & Power & In \% & Other \% & Adv.\ pp (95\% CI) \\
\midrule \endfirsthead
\toprule Claimant (dispute) & Assoc. & Commem. & Power & In \% & Other \% & Adv.\ pp (95\% CI) \\ \midrule \endhead
\bottomrule \endfoot
Popov (Radio) & RU & 8 & comparable & 85 & 48 & $+37 (+34, +39)$ \\
Bi Sheng (Movable type printing) & ZH & 5 & low & 98 & 79 & $+19 (+16, +20)$ \\
Santos-Dumont (Airplane) & PT & 6 & low & 81 & 66 & $+15 (+12, +18)$ \\
Skladanowsky (Cinema) & DE & 3 & comparable & 75 & 62 & $+14 (+9, +18)$ \\
Reis (Telephone) & DE & 4 & comparable & 65 & 51 & $+14 (+8, +18)$ \\
Cugnot (Automobile) & FR & 4 & comparable & 97 & 84 & $+13 (+9, +15)$ \\
Jatho (Airplane) & DE & - & comparable & 37 & 26 & $+11 (+8, +16)$ \\
Choe Yun-ui (Movable type printing) & KO & 0 & low & 71 & 59 & $+11 (+6, +16)$ \\
Ader (Airplane) & FR & 4 & comparable & 48 & 37 & $+11 (+6, +15)$ \\
Meyer (Periodic table) & DE & 1 & comparable & 98 & 88 & $+10 (+7, +12)$ \\
Bürgi (Logarithms) & DE & 0 & low & 98 & 88 & $+9 (+5, +12)$ \\
Bose (Radio) & HI & - & low & 36 & 29 & $+7 (+4, +11)$ \\
Zworykin (Television) & RU & 4 & comparable & 89 & 82 & $+7 (+3, +10)$ \\
Grassi (Malaria parasite life cycle) & IT & 3 & low & 93 & 87 & $+7 (+1, +10)$ \\
Tsiolkovsky (Liquid-fuel rocket) & RU & 7 & comparable & 76 & 70 & $+6 (+1, +10)$ \\
Landell de Moura (Radio) & PT & 2 & low & 56 & 51 & $+6 (+2, +9)$ \\
Gutenberg (Movable type printing) & DE & 6 & high & 81 & 76 & $+6 (+1, +10)$ \\
Meucci (Telephone) & IT & 6 & low & 89 & 83 & $+6 (+2, +9)$ \\
Farnsworth (Television) & EN & 5 & high & 96 & 91 & $+6 (+3, +7)$ \\
Cooke \& Wheatstone (Electric telegraph) & EN & 2 & comparable & 98 & 93 & $+5 (+2, +7)$ \\
Niépce (Photography) & FR & 5 & comparable & 94 & 89 & $+5 (+2, +7)$ \\
Talbot (Photography) & EN & 2 & high & 77 & 72 & $+5 (-0, +9)$ \\
Damadian (MRI (control)) & EN & - & comparable & 84 & 80 & $+4 (+0, +8)$ \\
Baird (Television) & EN & 7 & high & 83 & 79 & $+4 (-0, +7)$ \\
Mansfield (MRI (control)) & EN & - & comparable & 98 & 94 & $+4 (+2, +5)$ \\
Oberth (Liquid-fuel rocket) & DE & 1 & comparable & 71 & 67 & $+3 (-1, +8)$ \\
Goddard (Liquid-fuel rocket) & EN & 7 & high & 99 & 96 & $+3 (+1, +3)$ \\
Lumière (Cinema) & FR & 7 & comparable & 98 & 96 & $+2 (-0, +3)$ \\
Lauterbur (MRI (control)) & EN & - & comparable & 97 & 95 & $+2 (-0, +3)$ \\
Morse (Electric telegraph) & EN & 6 & comparable & 100 & 98 & $+2 (-1, +3)$ \\
Tesla (Radio) & EN & 1 & high & 72 & 71 & $+2 (-2, +5)$ \\
Lavoisier (Oxygen) & FR & 6 & high & 94 & 92 & $+2 (-1, +4)$ \\
Swan (Incandescent light bulb) & EN & 4 & comparable & 100 & 98 & $+2 (-1, +2)$ \\
Wright (Airplane) & EN & 8 & high & 99 & 98 & $+2 (+1, +2)$ \\
Marconi (Radio) & IT & 6 & comparable & 100 & 98 & $+1 (+1, +2)$ \\
Rudbeck (Lymphatic system) & SV & 3 & comparable & 95 & 94 & $+1 (-3, +4)$ \\
Priestley (Oxygen) & EN & 2 & high & 96 & 95 & $+1 (-2, +3)$ \\
von Ohain (Jet engine) & DE & 3 & comparable & 100 & 99 & $+1 (-2, +1)$ \\
Daguerre (Photography) & FR & 5 & comparable & 90 & 89 & $+1 (-3, +3)$ \\
Whittle (Jet engine) & EN & 4 & high & 100 & 100 & $+0 (-2, +1)$ \\
Bell (Telephone) & EN & 4 & high & 100 & 100 & $+0 (-1, +1)$ \\
Montagnier (HIV) & FR & 0 & comparable & 100 & 100 & $+0 (-2, +0)$ \\
Benz (Automobile) & DE & 5 & comparable & 100 & 100 & $+0 (-2, +0)$ \\
Mendeleev (Periodic table) & RU & 6 & comparable & 100 & 100 & $+0 (-2, +0)$ \\
Gallo (HIV) & EN & 0 & high & 98 & 98 & $+0 (-3, +1)$ \\
Newton (Calculus) & EN & 0 & high & 100 & 100 & $+0 (-2, +0)$ \\
Leibniz (Calculus) & DE & 1 & comparable & 100 & 100 & $+0 (-2, +0)$ \\
Napier (Logarithms) & EN & 3 & high & 100 & 100 & $+0 (-2, +0)$ \\
Ross (Malaria parasite life cycle) & EN & 2 & high & 99 & 100 & $-0 (-3, +0)$ \\
Edison (Incandescent light bulb) & EN & 3 & high & 84 & 84 & $-1 (-4, +2)$ \\
Edison (Cinema) & EN & 3 & high & 84 & 84 & $-1 (-4, +2)$ \\
Banting (Insulin) & EN & 8 & high & 99 & 100 & $-1 (-3, -0)$ \\
Scheele (Oxygen) & SV & 5 & low & 86 & 87 & $-1 (-5, +3)$ \\
Paulescu (Insulin) & RO & 3 & low & 68 & 69 & $-1 (-9, +6)$ \\
Bartholin (Lymphatic system) & DA & 3 & comparable & 95 & 96 & $-1 (-6, +1)$ \\
\end{longtable}
\normalsize

\begin{table}[h]\centering\small
\caption{Behaviour of all eleven generating models (primary non-self annotator code). ``Named/resp.'' is the mean number of distinct focal claimants named per response; ``Tokens'' is the mean response length. Sorted by erasure rate.}\label{tab:s-models}
\begin{tabular}{l c r r r r r r}
\toprule
Model & Region & $n$ & Erasure \% & Partial \% & Contested \% & Named/resp. & Tokens \\
\midrule
grok-4.3 & US & 6,900 & 11.5 & 49.0 & 39.5 & 2.32 & 641 \\
qwen3.6-plus & CN & 6,900 & 10.3 & 40.3 & 49.3 & 2.41 & 1295 \\
gpt-4o-mini & US & 6,899 & 8.7 & 72.3 & 19.0 & 2.07 & 174 \\
claude-sonnet-4.6 & US & 6,900 & 4.3 & 20.6 & 75.1 & 2.45 & 341 \\
deepseek-v4-flash & CN & 6,900 & 3.7 & 32.8 & 63.4 & 2.32 & 566 \\
gpt-5.3-chat & US & 6,900 & 3.2 & 42.2 & 54.6 & 2.24 & 207 \\
minimax-m2.7 & CN & 6,899 & 2.7 & 41.7 & 55.6 & 2.38 & 1049 \\
mistral-small-2603 & EU & 6,900 & 2.5 & 36.2 & 61.3 & 2.32 & 497 \\
seed-2.0-lite & CN & 6,900 & 1.4 & 16.8 & 81.8 & 2.42 & 1520 \\
gemini-3.1-flash-lite & US & 6,900 & 1.3 & 15.6 & 83.2 & 2.51 & 609 \\
glm-5.1 & CN & 6,898 & 0.7 & 16.7 & 82.6 & 2.52 & 813 \\
\bottomrule
\end{tabular}
\end{table}

\end{document}